\PassOptionsToPackage{unicode}{hyperref}
\PassOptionsToPackage{hyphens}{url}
\PassOptionsToPackage{dvipsnames,svgnames,x11names}{xcolor}
\documentclass[
  11pt,
  letterpaper,
]{article}
\usepackage{xcolor}
\usepackage[margin=0.65in]{geometry}
\usepackage{amsmath,amssymb}
\usepackage{iftex}
\ifPDFTeX
  \usepackage[T1]{fontenc}
  \usepackage[utf8]{inputenc}
  \usepackage{textcomp} 
\else 
  \usepackage{unicode-math} 
  \defaultfontfeatures{Scale=MatchLowercase}
  \defaultfontfeatures[\rmfamily]{Ligatures=TeX,Scale=1}
\fi
\usepackage{lmodern}
\ifPDFTeX\else
\fi
\IfFileExists{upquote.sty}{\usepackage{upquote}}{}
\IfFileExists{microtype.sty}{
  \usepackage[]{microtype}
  \UseMicrotypeSet[protrusion]{basicmath} 
}{}
\makeatletter
\@ifundefined{KOMAClassName}{
  \IfFileExists{parskip.sty}{%
    \usepackage{parskip}
  }{
    \setlength{\parindent}{0pt}
    \setlength{\parskip}{6pt plus 2pt minus 1pt}}
}{
  \KOMAoptions{parskip=half}}
\makeatother
\usepackage{longtable,booktabs,array}
\usepackage{calc} 
\usepackage{etoolbox}
\makeatletter
\patchcmd\longtable{\par}{\if@noskipsec\mbox{}\fi\par}{}{}
\makeatother
\IfFileExists{footnotehyper.sty}{\usepackage{footnotehyper}}{\usepackage{footnote}}
\makesavenoteenv{longtable}
\usepackage{graphicx}
\makeatletter
\newsavebox\pandoc@box
\newcommand*\pandocbounded[1]{
  \sbox\pandoc@box{#1}%
  \Gscale@div\@tempa{\textheight}{\dimexpr\ht\pandoc@box+\dp\pandoc@box\relax}%
  \Gscale@div\@tempb{\linewidth}{\wd\pandoc@box}%
  \ifdim\@tempb\p@<\@tempa\p@\let\@tempa\@tempb\fi
  \ifdim\@tempa\p@<\p@\scalebox{\@tempa}{\usebox\pandoc@box}%
  \else\usebox{\pandoc@box}%
  \fi%
}
\def\fps@figure{htbp}
\makeatother
\providecommand{\tightlist}{%
  \setlength{\itemsep}{0pt}\setlength{\parskip}{0pt}}
\usepackage{fancyhdr}
\AtBeginDocument{\linespread{0.97}\selectfont}
\AtBeginDocument{\renewcommand{\setminus}{\mathbin{\backslash}}}
\AtBeginDocument{\hypersetup{pdftitle={Can Semantic Geometry Teach an AI Judgement?}, pdfauthor={Thomson D. Nguy}, pdfsubject={Four bounded studies of semantic measurements for pre-action judgment}}}
\usepackage{bookmark}
\IfFileExists{xurl.sty}{\usepackage{xurl}}{} 
\hypersetup{
  colorlinks=true,
  linkcolor={black},
  filecolor={Maroon},
  citecolor={Blue},
  urlcolor={blue},
  pdfcreator={LaTeX via pandoc}}

\author{}
\date{}

\begin{document}

\begin{center}\LARGE\bfseries Can Semantic Geometry Teach an AI Judgement?\par\vspace{0.5em}\normalsize\normalfont Thomson D. Nguy\par\small Radiant Institute for Manifold Studies\end{center}

\subsection{Abstract}\label{abstract}

How can an AI agent determine what rules to follow? One rule permits an
action. Another imposes a condition, exception, or conflicting
obligation. Deterministic systems can resolve those relationships when
they have been specified. When they remain implicit in language, an
agent can follow one rule while missing another that should stop it.
Refusing every unresolved action avoids that risk, but also blocks
permissible actions.

We wanted the agent to make the distinction and still act. Our initial
hypothesis was that geometric measurements could supply a basis for
judgment. We represented actions and policies as vectors, then tested
whether their geometry could identify governing policies and interpret
the action's relation to them.

Across four studies, the tested approaches did not establish reliable
pre-action judgment. In the final synthetic study, a lexical router
recovered every governing and blocking policy while reducing median
policy checks by 97.7\%. The composed pipeline nevertheless escalated
all 2,304 test actions, including those it should have allowed.
Supplying every policy to the same downstream mechanism changed no
decision. Finding the policies had not solved the problem of
interpreting them.

This result led us to revise our hypothesis: judgment in AI agents
requires developing a consequence graph. Such a graph would connect the
actor and authority to policy conditions, exceptions, and the changes an
action would produce. Follow-on studies will ask whether making those
relationships explicit helps the agent distinguish when to act, stop, or
seek review.

\begin{minipage}{\linewidth}

\includegraphics[width=1\linewidth,height=\textheight,keepaspectratio,alt={Figure 1. From a proposed action to a governed decision.}]{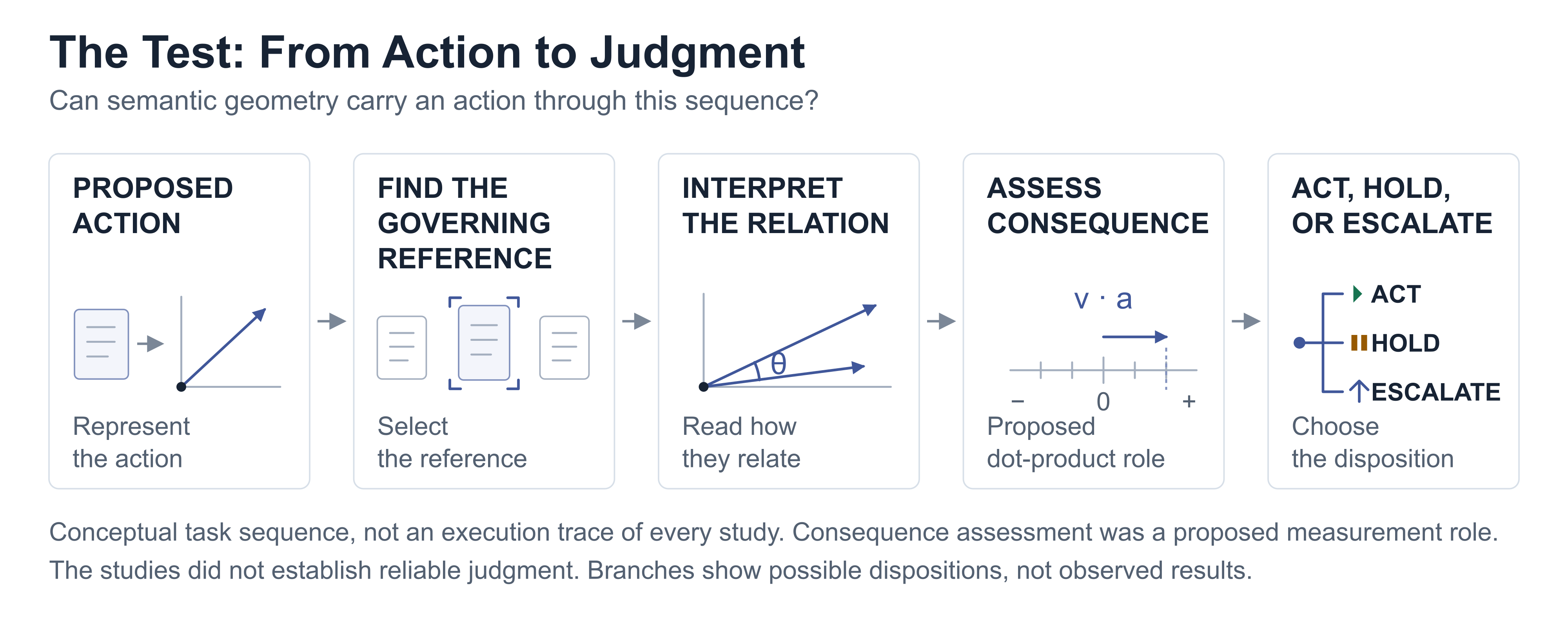}

\emph{Figure 1. The sequence the original hypothesis asked geometry to
support. Consequence assessment was a proposed role for signed raw dot
product. The studies did not establish that role or reliable pre-action
judgment.}

\end{minipage}\par\medskip

\subsection{1. Governed autonomy}\label{governed-autonomy}

\subsubsection{Several policies can govern the same
action}\label{several-policies-can-govern-the-same-action}

Consider an engineering agent asked to move its owner's API credential
from an owner-controlled environment into an approved secrets manager
using access the owner has already authorized.

In this illustration, an authorization policy permits the agent to act
on the owner's behalf. A handling policy requires the credential to go
to an approved secrets manager. A confidentiality policy prohibits
giving it to an unauthorized recipient. Permission under the first
policy has to be considered with the conditions imposed by the others.

Now change the destination. Instead of moving the credential into the
approved secrets manager, the agent sends it to an unauthorized external
party. Much of the language remains the same. The actor may be the same.
The object is the same. The general action still involves moving a
credential from one place to another. The governance consequence is
different because the destination, recipient, authority, and resulting
state are different.

Formal policy systems already perform versions of this work when
policies, targets, and facts have been expressed in structured form
{[}1, 2{]}. Runtime shields can enforce explicit constraints against
modeled states {[}5{]}. Their strength comes from making the governing
relationships explicit. Their limit is that the policies, targets, and
relevant state must already be represented in a form the evaluator can
use.

If those relationships have not been made explicit, a rule that refuses
every credential transfer avoids the dangerous action. It also blocks
the authorized one. The system no longer has to distinguish between
them.

An autonomous agent presents the harder problem. It has to distinguish
actions that look similar in language but differ in the relationships
that make them permitted, prohibited, conditional, or unresolved.

We wanted to preserve safety while letting the agent make this
distinction and still act.

\subsubsection{What the human decision-maker
reconstructs}\label{what-the-human-decision-maker-reconstructs}

An experienced decision-maker does not compare the wording of the
proposed action with the wording of the policy and stop at semantic
proximity. The decision-maker reconstructs the event.

Who is acting? In what role? Under whose authority? What action is being
taken, and on what object? Where is the object moving from, and where is
it going? Who receives the effect? What condition is present? Does an
exception apply? What state exists now, and what state will exist if the
action completes?

The decision-maker can still be wrong. But the unit of analysis is a
structure of relationships, not the similarity between two
undifferentiated blocks of text.

A policy has a similar internal structure. It may identify an actor or
class of actors, an action, an object, a condition, an exception, an
authority, and a consequence. The policy becomes relevant when the
contemplated event enters its semantic neighborhood. It activates only
when the event instantiates the relationships under which the policy
operates. A consequence follows only after that activation is resolved.

The unresolved problem appears earlier. Many proposed actions arrive as
natural language. The relevant policy may be buried among hundreds of
rules, objectives, and higher-order missions. Some governing facts may
be expressed indirectly. Before a structured evaluator can operate, the
system has to find the policy and reconstruct enough of the event to
determine whether that policy activates.

Finding the policy and determining that it activates are different
problems.

\subsubsection{The measurement wager}\label{the-measurement-wager}

The original proposal tried to cross that gap with semantic geometry. It
represented the proposed action and the governing references as vectors,
then compared them before execution.

The attraction was real. A vector system could work across variations in
language that a literal rule might miss, search a large policy field
quickly, and perhaps provide graded signals rather than force every
unfamiliar action into an immediate refusal.

The conception assigned different roles to three measurements. Raw
Euclidean distance might help locate the references nearest to the
proposed action. Cosine similarity might indicate whether the action
moved with or against the direction of a governing reference. Signed raw
dot product might carry information about the magnitude of the
consequence. A separate rule would then choose ACT (proceed), HOLD (do
not act), or ESCALATE (send for review).

These governing references include concrete rules, operational
objectives, and explicit higher-order missions. We use judgment here for
the proposed ability to interpret an action's relation to those
references and distinguish when to act, stop, or seek review.

The sequence in Figure 1 gave the measurements a job. It did not
establish that they could perform it.

Once the action becomes one vector, the actor, authority, object,
destination, condition, exception, and anticipated state are no longer
presented to the decision mechanism as explicit relationships. Once the
policy becomes another vector, its activation conditions and consequence
structure are flattened as well. The distance between the vectors may
preserve semantic information. Whether it preserves the relationships
needed for governed action is an empirical question.

\subsubsection{The first warning}\label{the-first-warning}

Even before running that experiment, the mathematics exposed a narrower
problem. Three named formulas do not necessarily produce three
independent measurements.

Let the raw proposed-action vector be \(v \in \mathbb{R}^d\) and the raw
reference vector be \(a \in \mathbb{R}^d\), both finite and nonzero.
Under total L2 normalization,

\[
u = \frac{v}{\lVert v \rVert_2}, \qquad r = \frac{a}{\lVert a \rVert_2},
\]

with \(\lVert u \rVert_2 = \lVert r \rVert_2 = 1\). Let
\(c = u^\top r = \cos\theta\). For the specified normalized components,

\[
D_C = 1-c, \qquad D_E^2 = \lVert u-r\rVert_2^2 = 2(1-c), \qquad D_P = 1-u^\top r = 1-c.
\]

The weighted form

\[
R_{\mathrm{raw}} = \alpha D_C + \beta D_E^2 + \gamma D_P
\]

therefore becomes

\[
R_{\mathrm{raw}} = (\alpha + 2\beta + \gamma)(1-c).
\]

With nonnegative weights and at least one positive weight, the
combination preserves the cosine-drift ordering. The three formulas have
collapsed to one angular ordering. A later monotone normalization can
change the scale. It cannot restore information already removed.

The derivation applies to these formulas after total L2 normalization.
It does not cover every metric, encoder, normalization, or alternative
design. Before normalization, vector norms remain available:

\[
v^\top a = \lVert v\rVert_2\lVert a\rVert_2 c,
\]

\[
\lVert v-a\rVert_2^2 = \lVert v\rVert_2^2 + \lVert a\rVert_2^2 - 2\lVert v\rVert_2\lVert a\rVert_2c.
\]

Those raw quantities preserve information that separate unit
normalization discards. They still have to earn the meanings assigned to
them. A distance does not become policy relevance because it is small.
An angle does not become relational direction because its sign looks
suggestive. A dot product does not become consequence because its
magnitude is large.

The proposal therefore faced two questions. Were the measurements
mathematically distinct under the formulation being used? And when
distinct information remained, did it capture the relational work
required for governed autonomy?

We tested those questions four times.

The studies did not test whether a graph can reproduce senior human
judgment. They did not establish that any particular structured
successor is necessary or sufficient. They tested the vector lineage on
its own terms. The successor question comes later, after the failure has
been located.

\subsection{2. We tested it four times}\label{we-tested-it-four-times}

We did not begin with four studies. We began with one representation and
one question.

Could a compact numerical description preserve enough information for a
fixed reader to distinguish the ways an action might relate to a
governing reference?

The answer to that question did not settle the program. It changed the
next question.

\subsubsection{The first test: could the compact representation recover
the missing
distinctions?}\label{the-first-test-could-the-compact-representation-recover-the-missing-distinctions}

The first study tested a normalized scalar augmented by the norms of the
action and reference vectors. The idea was straightforward.
Normalization removes magnitude information. If the scalar relation and
the two original norms were supplied together, perhaps a fixed reader
could recover joint information that the scalar alone had lost.

The fixed corpus contained 720 canonical action-reference pairs, each
rendered in three controlled surface forms (2,160 blinded rows total).
Each of the three encoders was evaluated on the same rendered set. The
reader had to separate four cells. This was a deliberately bounded test
of one representation and one reader, not a test of every possible
metric or model {[}6{]}.

For encoders E01, E02, and E03, the primary gains were
\texttt{0.010416666667}, \texttt{0.062500000000}, and
\texttt{0.020833333333}, respectively. All were below the prespecified
\texttt{0.10} threshold, and none passed the full primary gate. The
study did not find stable separable signal on its benchmark.

That result closed the broad claim for the exact benchmark. The
scalar-plus-norm profile had not produced stable four-cell separation.

It did not tell us that every quantity inside the profile was empty. A
joint reader can fail even when one input carries a narrower signal. The
next study therefore pulled the measurements apart.

\subsubsection{The second test: did the measurements mean what their
names
suggested?}\label{the-second-test-did-the-measurements-mean-what-their-names-suggested}

The second study tested cosine similarity, raw Euclidean distance, and
signed raw dot product as separate features before testing them in
combination. The fixed panel contained 936 canonical action-reference
pairs across twelve families. Each pair was rendered in three controlled
surface forms (2,808 blinded rows total), and each of the three encoders
evaluated that full rendered set. The row-level split comprised 936
calibration and 1,872 sealed rows per encoder. Seven isolated and
combined arms allowed each proposed role to answer for itself {[}7{]}.

This time, the result was mixed.

Raw Euclidean distance was workable for proximity in the exact panel.
Cosine carried some directional information, although the result
narrowed for one encoder. Signed raw dot product was misleading as a
measure of consequence magnitude. The full three-output profile did not
Pareto-dominate the alternatives and received a \texttt{NULL} terminal.

The original three-part story had not survived. But neither had the
simpler conclusion that nothing in the geometry worked. Two bounded
component signals remained: proximity and a narrowed form of direction.

That created the next question. Perhaps the full profile did not need to
win as one undifferentiated score. Perhaps the surviving signals could
perform different jobs in sequence.

Euclidean distance could narrow the policies that deserved attention.
Cosine could then characterize the action's relation to the smaller
candidate set. A separate rule could convert those relations into ACT,
HOLD, or ESCALATE.

The third study tested that composition.

\subsubsection{The third test: could the surviving signals work
together?}\label{the-third-test-could-the-surviving-signals-work-together}

The fixed corpus contained 1,296 action cases and 10,368 action-policy
pair rows, generated across twelve families and split into 3,456
calibration and 6,912 sealed pair rows. Each of the three encoders was
evaluated on the same corpus under the eight prespecified arms. It asked
whether Euclidean routing followed by cosine relation mapping could
reduce policy review while preserving the policies that governed the
action, then produce accurate final dispositions {[}8{]}.

The routing results looked promising. Depending on the encoder, the
system retained \texttt{96.8\%} to \texttt{99.8\%} of governing policies
while reducing policy evaluation by \texttt{52.1\%} to \texttt{57.8\%}.

Then the pipeline had to decide what to do.

Final-disposition accuracy ranged from \texttt{44.6\%} to
\texttt{58.9\%}. The broader composed claim was \texttt{NULL} in every
encoder, with \texttt{0/8} family support. Geometry had reduced the
amount of policy material the system inspected. That reduction had not
become reliable ACT, HOLD, or ESCALATE decisions.

The result left several possibilities entangled. The router had missed
some governing policies. The relation stage might have misread policies
that were present. The action rule might have destroyed discriminating
relation information during aggregation. More than one stage could have
contributed.

The third study showed that the pipeline had not worked. It could not
tell us which part had limited it.

The fourth study was built to hold those parts apart.

\subsubsection{The fourth test: make each stage answer
separately}\label{the-fourth-test-make-each-stage-answer-separately}

The fourth study began again in a fresh synthetic policy world. The
fixed corpus began with 720 base actions and 40,320 base action-policy
pairs. Each base action had seven prespecified perturbations, yielding
5,760 action records and 322,560 action-policy pair rows across thirty
disjoint mission families. Each of the two fixed encoders was evaluated
on the same expanded corpus; calibration, validation, and sealed pair
rows were 129,024, 64,512, and 129,024, respectively. The study froze
separate tests for routing, relation classification, and end-to-end
composition before sealed scoring {[}9{]}.

The world included actions with no governing policy, one governing
policy, several compatible governing policies, and several conflicting
governing policies. The test therefore included cases where the system
had to interpret policies that applied together, not just find a single
relevant rule {[}9{]}.

The routing experiment compared four isolated routers and four control
families. Its question was narrow: could a router reduce the policies
sent forward for review without dropping the governing or blocking
policies?

The relation experiment removed retrieval scarcity from the problem. An
oracle defined a separately sampled test set containing governing
policies and explicit non-governing controls, independent of router
selection. Seven isolated and combined feature arms then tested whether
cosine, raw Euclidean distance, and signed raw dot product improved
relation classification, separately or together.

The composition experiment put the selected router and the full
three-measure relation model back into one pipeline. It compared that
candidate with exhaustive semantic evaluation, the prior two-stage
baseline, isolated measurement routes, top-k controls, and a
deterministic oracle.

The study found three different outcomes:

\begin{itemize}
\tightlist
\item
  The selected router reduced policy review without missing a governing
  or blocking policy on the sealed surface.
\item
  The additional measurements did not materially improve relation
  classification over the best single measurement.
\item
  The complete pipeline did not produce reliable ACT, HOLD, and ESCALATE
  decisions.
\end{itemize}

Those results complete the four-study sequence at program scale. The
first study found no stable joint information in its compact reader. The
second preserved bounded proximity and direction signals while rejecting
magnitude and full-profile claims. The third showed that the surviving
signals did not automatically compose into accurate dispositions. The
fourth separated the stages and made their different outcomes
observable.

\subsection{3. The fourth test makes the failure
legible}\label{the-fourth-test-makes-the-failure-legible}

Finding the policy was not the problem. The selected router recovered
every sealed governing and blocking policy while cutting median policy
checks by 97.7 percent. The composed system still escalated every
action.

Those results came from the same frozen study. By separating routing,
relation interpretation, and final disposition, the fourth study
prevented one stage from hiding behind another. Each stage had to answer
for itself {[}9{]}.

\subsubsection{Follow one action through the
test}\label{follow-one-action-through-the-test}

Let \(x\) be a proposed action and \(P_x\) the policies available for
that action. The router returns a candidate set \(S_x = R(x,P_x)\),
where \(S_x \subseteq P_x\). For scoring, the synthetic world's oracle
supplies \(G_x\), the governing policies, and \(B_x \subseteq G_x\), the
blocking policies. The system being tested does not get those oracle
sets when it routes an action.

The routing question is whether \(S_x\) keeps the policies the oracle
says matter while sending fewer policies to the next stage. Across
sealed actions, governing-policy recall is

\[
\frac{\sum_x |S_x \cap G_x|}{\sum_x |G_x|}.
\]

For each action, the fraction of policy checks saved is
\(1-|S_x|/|P_x|\). We report the median of that fraction across actions.
Blocking misses count oracle blocking instances outside the candidate
set, \(\sum_x |B_x\setminus S_x|\). A small candidate set is no
achievement if it drops the rule that would stop the action.

The next stage receives each tested action-reference pair associated
with a policy \(p \in S_x\). Let \(v_e(x)\) be the raw action vector and
\(a_e(x,p)\) its raw reference vector under fixed encoder \(e\). In the
full measurement arm, the relation reader receives

\[
z_e(x,p)=\left(\cos(v_e(x),a_e(x,p)),\;\lVert v_e(x)-a_e(x,p)\rVert_2,\;v_e(x)^\top a_e(x,p)\right).
\]

A frozen multinomial logistic reader converts those three values into
probabilities for ALIGNED, OPPOSED, AMBIGUOUS, and NON\_GOVERNING. The
fixed aggregation rule then returns an action disposition:

\[
S_x=R(x,P_x)
\quad\longrightarrow\quad
q_e(\ell\mid x,p),\;p\in S_x
\quad\longrightarrow\quad
\hat y_e(x)\in\{\mathrm{ACT},\mathrm{HOLD},\mathrm{ESCALATE}\}.
\]

Here \(e\) identifies one of the two fixed encoders and \(q_e\) is the
reader's probability for relation label \(\ell\). The selected-model
probability threshold was 0.70. The aggregation rule discarded
predictions labeled NON\_GOVERNING and sent the action to ESCALATE if
none remained or any retained prediction was AMBIGUOUS or had a maximum
class probability below 0.70; otherwise, any retained OPPOSED prediction
sent it to HOLD, and the remaining cases yielded ACT. This describes the
frozen experiment, not a general policy-decision algorithm. It also
makes the diagnostic separable: the router can find \(G_x\), the reader
can still misclassify a tested pair, and aggregation can still turn
those classifications into a poor action decision.

\begin{minipage}{\linewidth}

\includegraphics[width=1\linewidth,height=\textheight,keepaspectratio,alt={Figure 2. The fourth study separated reference selection, relation interpretation, and action disposition.}]{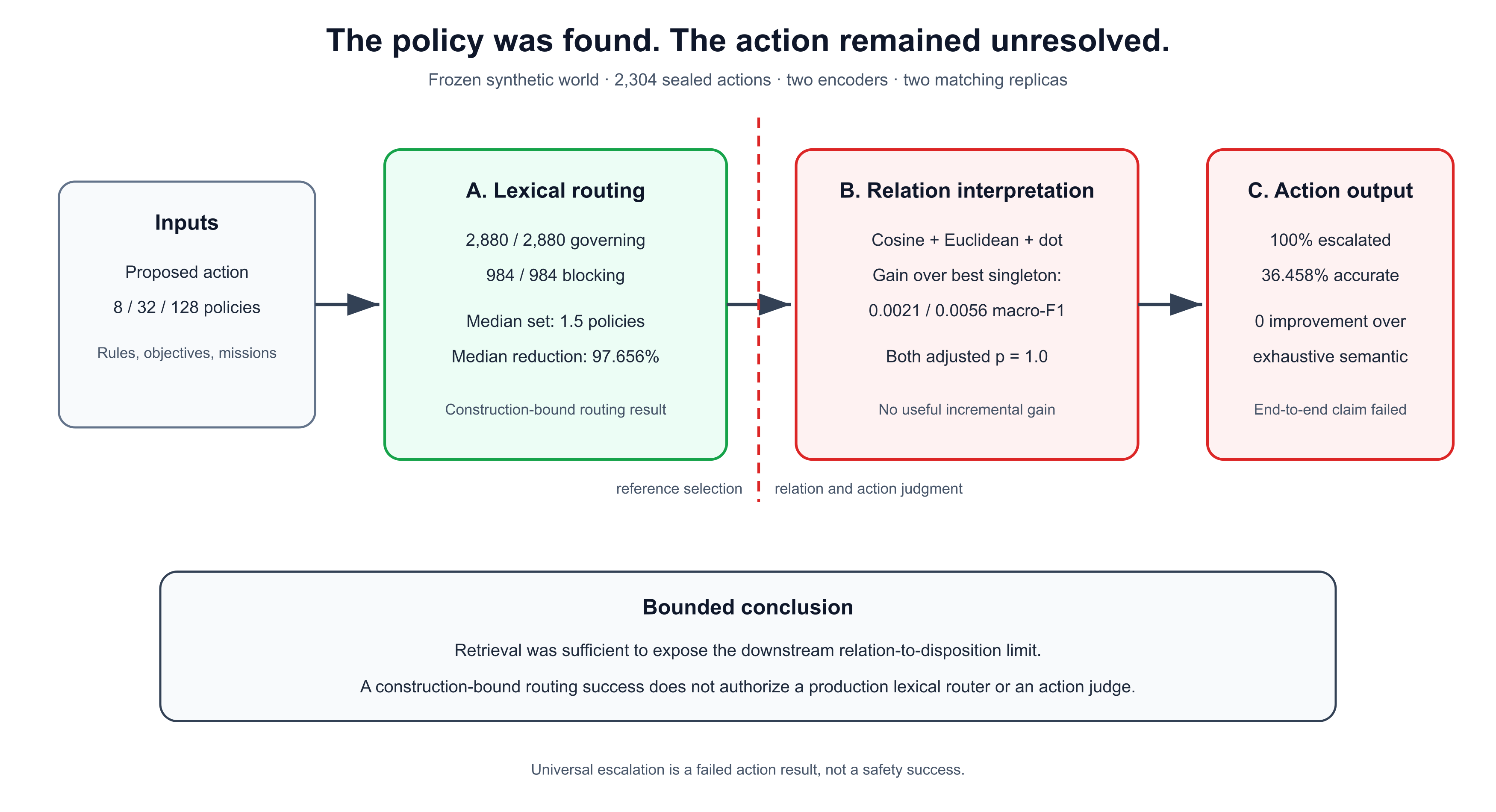}

\emph{Figure 2. The evaluated pipeline. On this constructed surface,
reference selection succeeded; relation interpretation and final
disposition did not. The figure reports separate outcomes, not a causal
proof that one feature or one rule alone produced universal escalation.}

\end{minipage}\par\medskip

\subsubsection{The policies were found}\label{the-policies-were-found}

The winning router was not the three-measure semantic system. It was a
lexical-overlap rule with fixed token and negation preprocessing.

On the sealed surface, that simple rule found \texttt{2,880/2,880}
governing-policy instances. It also found all \texttt{984/984}
blocking-policy instances. There were no blocking misses. The router
returned a median of \texttt{1.5} policies per action from policy sets
containing 8, 32, or 128 policies. Median policy evaluation fell by
exactly \texttt{0.976562500000}. False-retrieval burden was zero, and
the candidate set remained exactly stable across the frozen
perturbations.

Inside this policy world, the routing result was strong. It was also
construction-bound. The synthetic corpus exposed frozen route terms to
the pipeline, and the lexical rule operated on authored language with
fixed preprocessing. The result does not tell us that lexical overlap
will recover policies expressed through implication, unfamiliar
paraphrase, adversarial language, legal construction, or the accumulated
ambiguities of a real organization.

But it does tell us what happened here. On this surface, the policies
needed for judgment were present. The selected router found them.

\begin{minipage}{\linewidth}

\includegraphics[width=1\linewidth,height=\textheight,keepaspectratio,alt={Figure 3. Governing-policy recall and median reduction in policy checks by router.}]{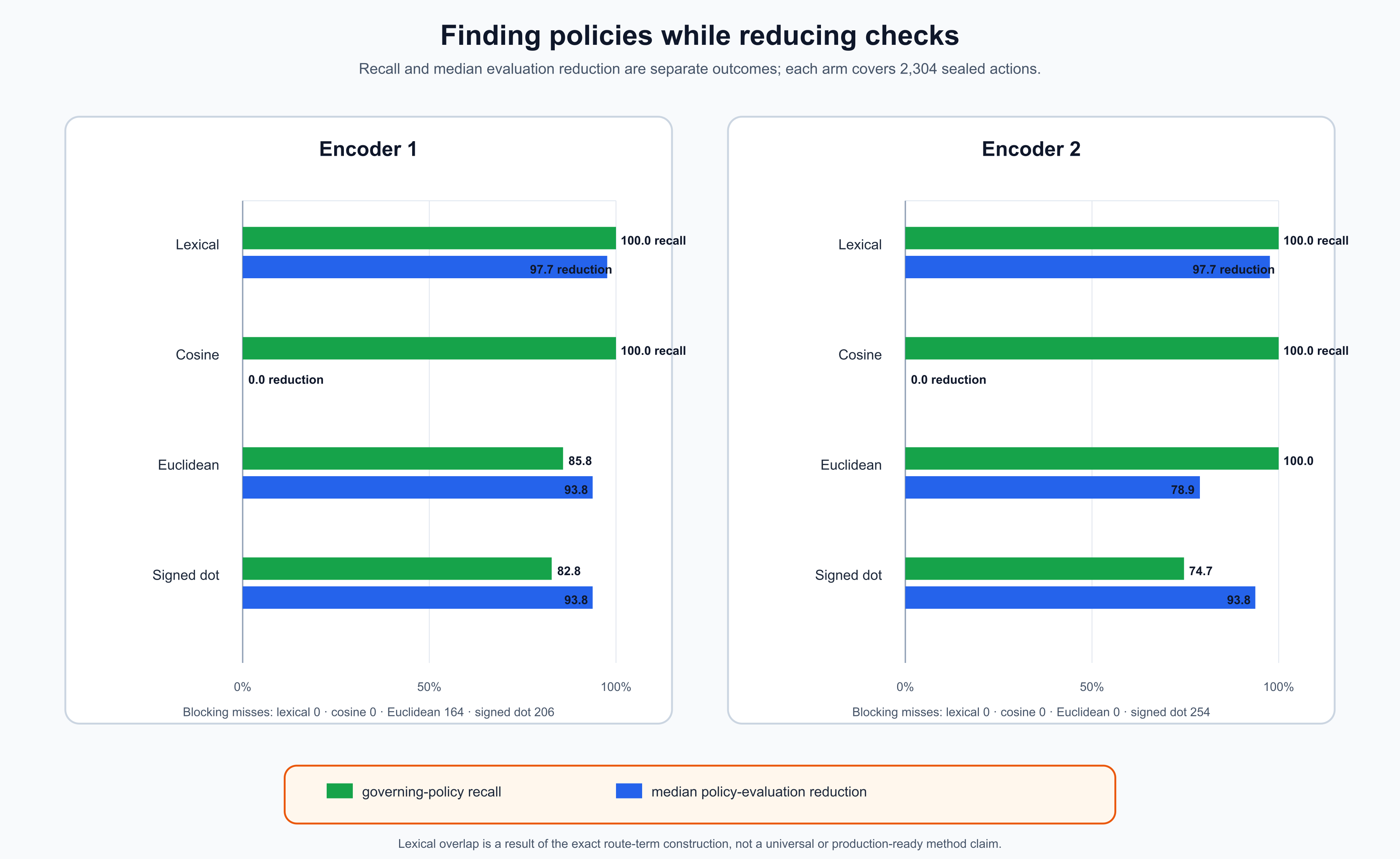}

\emph{Figure 3. Routing had two jobs: retain governing policies and
reduce checks. Cosine retained the former only by keeping a median
32-policy set; the lexical router achieved both on the authored
route-term surface. Blocking misses remain visible because ordinary
recall alone can conceal a dangerous omission.}

\end{minipage}\par\medskip

That removed one possible explanation for the failed pipeline. It did
not tell us whether the system understood the relation between an action
and the policies it had found.

\subsubsection{The measurements did not add the expected relation
signal}\label{the-measurements-did-not-add-the-expected-relation-signal}

The relation experiment took routing out of the way. After measurements
were sealed, an oracle supplied a separately sampled set containing
governing policies and explicit non-governing controls. Each arm saw
3,840 sealed action-policy pairs per encoder, independently of router
selection. No router chose which policies entered this test.

Seven isolated and combined feature arms then tested cosine similarity,
raw Euclidean distance, and signed raw dot product. The full
three-measure model reached relation macro-F1 of \texttt{0.355163305089}
with the first encoder and \texttt{0.485257556129} with the second. In
both encoders, Euclidean-only was the best singleton.

The primary comparison was paired within each of the twelve sealed
mission families. For encoder \(e\) and family \(f\), let
\(\Delta_{e,f}\) be that family's relation macro-F1 for all three
measurements minus its macro-F1 for Euclidean-only. The reported gain is
the mean of those twelve paired differences, not a subtraction of the
two corpus-level macro-F1 scores. Adding cosine and signed dot product
produced mean gains of \texttt{0.002145935796} with the first encoder
and \texttt{0.005580263897} with the second. The prespecified
useful-effect floor was \texttt{0.03}. The family-cluster bootstrap
intervals were \texttt{{[}-0.006380732317,\ 0.012731300481{]}} and
\texttt{{[}-0.011429689965,\ 0.023414950180{]}}. Both intervals crossed
zero. After Holm adjustment, both permutation values were
\texttt{1.000000000000}.

\begin{minipage}{\linewidth}

\includegraphics[width=1\linewidth,height=\textheight,keepaspectratio,alt={Figure 4. Relation macro-F1 across seven feature arms and two encoders.}]{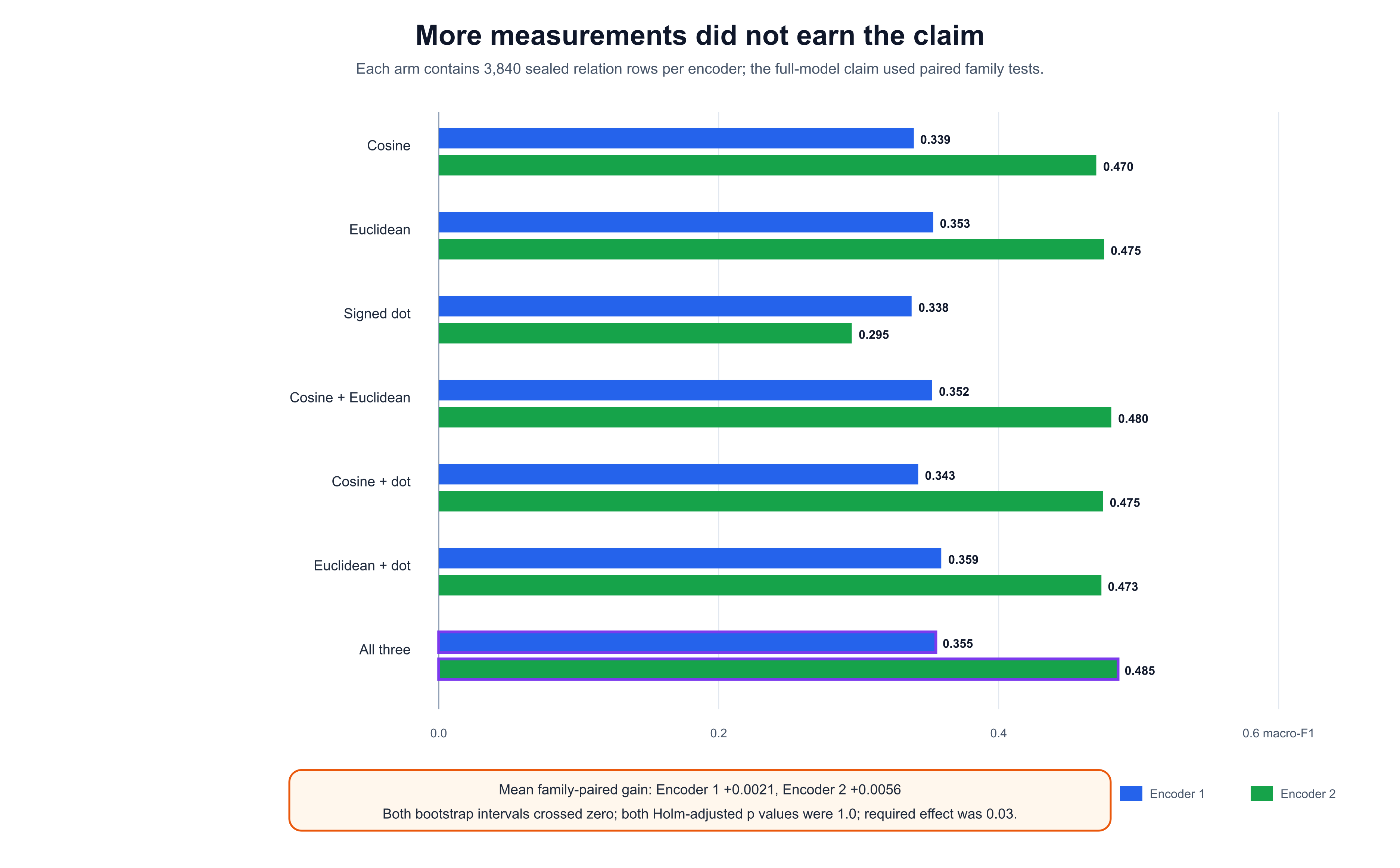}

\emph{Figure 4. The bars show descriptive macro-F1 for the seven arms.
The test of added utility used paired family differences against the
best singleton and the prespecified 0.03 floor; visual bar height alone
cannot establish that result.}

\end{minipage}\par\medskip

The question was not whether the full model emitted numbers. It did. The
question was whether the additional measurements improved relation
classification enough to justify the three-measure claim. They did not
on this frozen surface, in either encoder.

That result does not dispose of every possible relation model. The
fourth study tested two fixed encoders, one frozen model class, one
confidence threshold, and one synthetic world. It does dispose of the
claim that this full three-measure model had demonstrated incremental
relation discrimination clearing its prespecified floor over the best
singleton.

\subsubsection{Every action became
ESCALATE}\label{every-action-became-escalate}

The final test put the selected router and the full relation model back
into one pipeline. Then the pipeline had to decide among ACT, HOLD, and
ESCALATE.

It chose ESCALATE for all \texttt{2,304} sealed actions in each encoder.

The oracle expected \texttt{840} of those actions to be escalated. The
pipeline got those cases right. It also escalated all \texttt{672}
actions whose oracle disposition was ACT and all \texttt{792} actions
whose oracle disposition was HOLD. Final-disposition accuracy was
therefore \texttt{840/2,304}, or \texttt{0.364583333333}.

The system had zero false ACT decisions. That number looks favorable
until one asks how many ACT decisions the system made. The answer was
zero. The system avoided false permission by refusing to distinguish
permission from uncertainty.

\begin{minipage}{\linewidth}

\includegraphics[width=1\linewidth,height=\textheight,keepaspectratio,alt={Figure 5. Every sealed action was assigned ESCALATE.}]{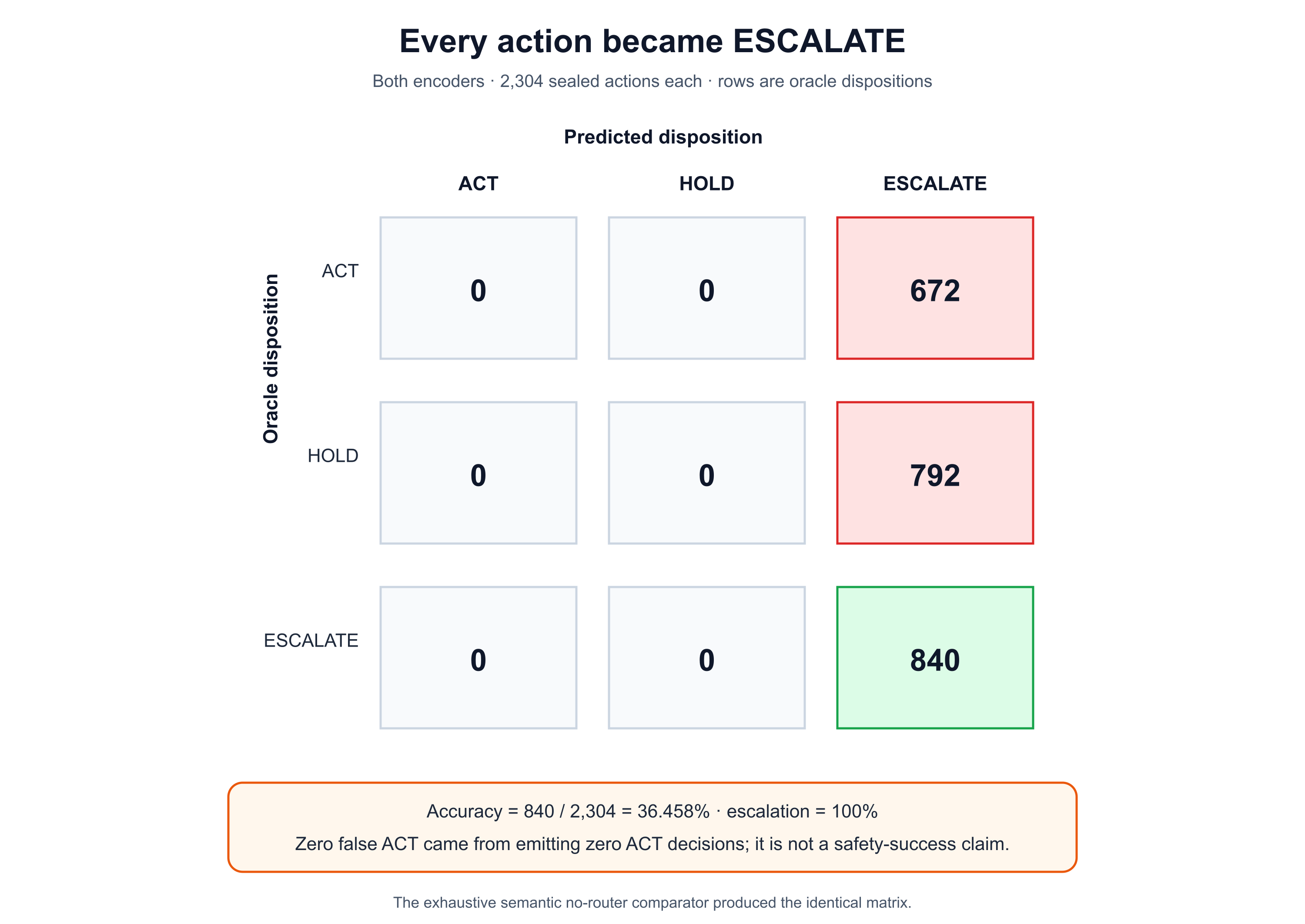}

\emph{Figure 5. The composed candidate's action-disposition matrix was
identical for both encoders: 672 oracle ACT actions and 792 oracle HOLD
actions became ESCALATE, while 840 oracle ESCALATE actions remained
ESCALATE. Zero false ACT reflects zero ACT predictions, not demonstrated
safety.}

\end{minipage}\par\medskip

Universal escalation can be appropriate in a world where every proposed
action is genuinely uncertain and review is free. That was not this test
world. ACT, HOLD, and ESCALATE cases were all present. A system that
converts every one of them into ESCALATE has not learned to make the
distinction the experiment asked it to make.

The end-to-end claim failed in both encoders.

\subsubsection{Give it every policy}\label{give-it-every-policy}

One possibility still remained. Perhaps the lexical router had recovered
every governing and blocking policy but had nevertheless changed the
policy context in a way that damaged the downstream decision.

The exhaustive semantic comparator removed that possibility. It sent
every policy forward while keeping the same relation model and
aggregation rule.

Giving the relation stage every policy did not change a single decision.

The exhaustive comparator escalated the same \texttt{2,304} actions. It
produced the same \texttt{0.364583333333} accuracy and the same
\texttt{1.000000000000} escalation rate. The mean accuracy difference
was zero in both encoders. Both bootstrap intervals were exactly
\texttt{{[}0.000000000000,\ 0.000000000000{]}}. Both permutation values
were \texttt{1.000000000000}.

The selected router had reduced the policy set sharply. Exhaustive
evaluation restored everything it removed. The result did not move.

That is what the fourth study made legible. In this exact frozen
mechanism, retrieval scarcity was not the dominant limit. The policies
were available. The downstream relation-to-disposition stage still did
not distinguish ACT, HOLD, and ESCALATE actions.

This is a localization, not yet an explanation. The experiment does not
tell us which alternative representation, relation model, threshold, or
aggregation rule would work. It tells us that adding more policies to
this downstream mechanism was not the repair.

\begin{minipage}{\linewidth}

\includegraphics[width=1\linewidth,height=\textheight,keepaspectratio,alt={Figure 6. What the four studies changed.}]{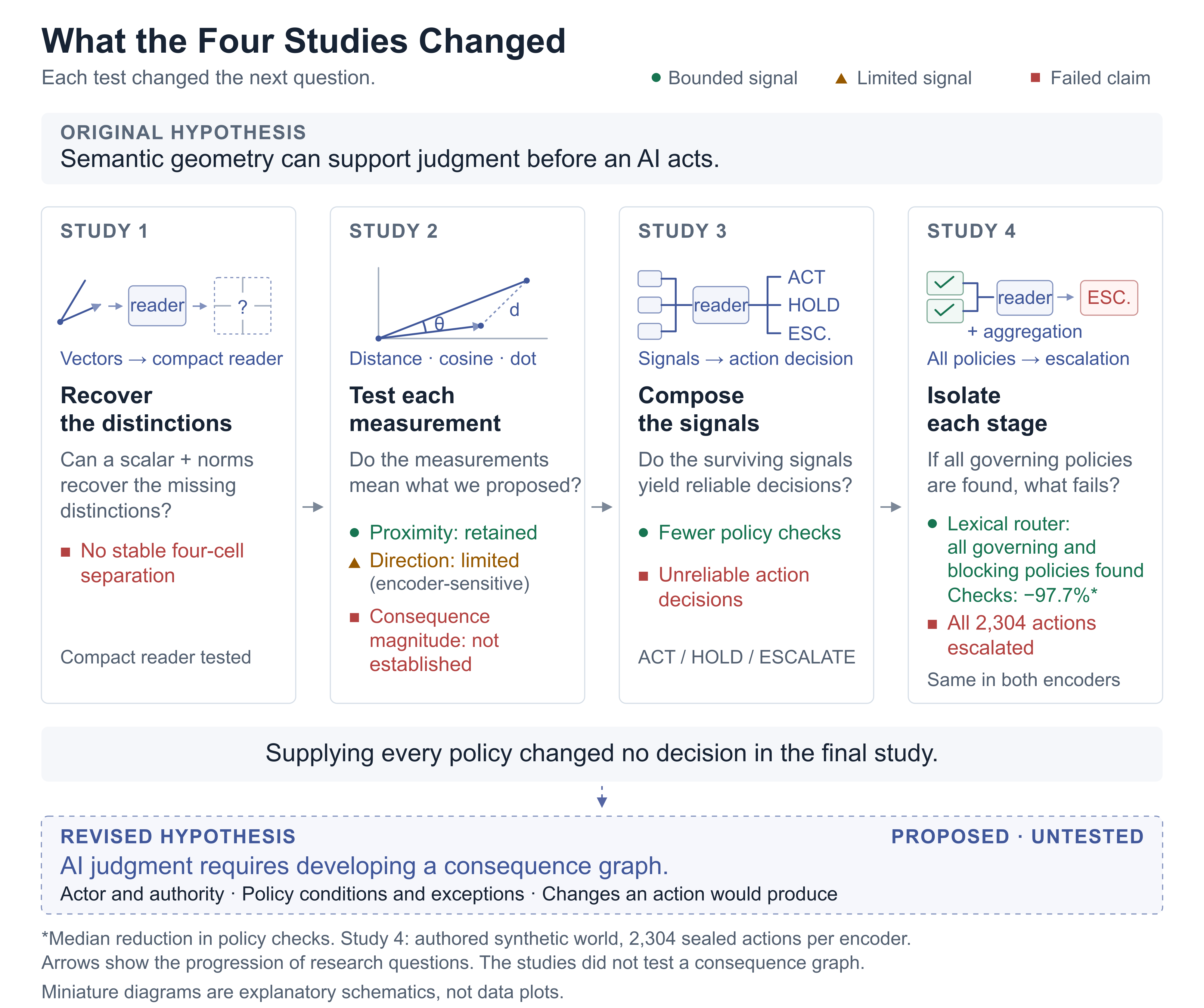}

\emph{Figure 6. Each test changed the next question. Bounded component
signals survived while the composed judgment claim failed. The
consequence-graph hypothesis is a proposed next step; none of the four
studies tested it.}

\end{minipage}\par\medskip

\subsection{4. What failed, what survived, and what remains
open}\label{what-failed-what-survived-and-what-remains-open}

\subsubsection{What failed}\label{what-failed}

The original conception gave three measurements a large job. Cosine
similarity, Euclidean distance, and signed dot product were supposed to
help distinguish whether a proposed action aligned with, departed from,
or opposed an explicit reference. If those measurements carried
independent information, and if their roles survived contact with data,
perhaps they could support a reliable pre-action judgment.

The mathematical derivation narrowed that claim before an empirical
model ever ran. Under the specified normalization, the three terms did
not provide three independent orderings. The first study then tested a
scalar-plus-norm representation and did not find stable recovery across
its frozen cells. The second returned to the raw quantities. Euclidean
proximity survived on its exact panel. Cosine retained a narrower,
encoder-sensitive directional role. Signed raw dot product did not
reliably order consequence severity, and the full three-output profile
did not establish itself as the better architecture.

The third study asked whether the surviving component roles could work
together. They did not compose into reliable ACT, HOLD, and ESCALATE
decisions. The fourth then gave the lineage its cleanest test. Retrieval
was separated from relation interpretation. The relation experiment
received a separately sampled set of governing policies and explicit
non-governing controls, and the full feature set was compared with its
best singleton. The additional measurements did not clear the
prespecified useful-effect floor. When routing and relation
interpretation were recombined, every action became ESCALATE.

The broad claim does not survive those four tests. On the frozen
corpora, encoders, readers, thresholds, and aggregation rules, the three
measurements did not establish themselves as a discriminating pre-action
judge.

The studies did not test every semantic policy architecture. They tested
this measurement lineage, then made it progressively harder for one
failed component to hide behind another.

\subsubsection{What survived}\label{what-survived}

The negative result did not erase every instrument. Raw Euclidean
distance retained bounded proximity information in the second study.
Cosine retained a narrower directional signal, although one encoder
required the claim to be narrowed. The third study showed that routing
could reduce the number of policy checks while retaining most governing
policies on its exact surface.

The fourth study produced the strongest positive result. Its selected
lexical router recovered all \texttt{2,880/2,880} sealed
governing-policy instances and all \texttt{984/984} blocking-policy
instances while reducing median policy evaluation by \texttt{97.7}
percent. That result belongs to an authored synthetic world with
explicit route terms and fixed preprocessing. It cannot be promoted into
a general claim about natural-language policy retrieval. Inside that
world, however, it did what the test asked it to do. It found the
policies and sharply reduced the review set.

The surviving result is therefore more than a consolation prize, but
less than an action judge. The system could allocate attention. It could
not yet determine whether the action satisfied, violated, triggered, or
remained uncertain relative to the policies it had found.

That division of labor became clearer with each study. Policy
localization, relation interpretation, uncertainty, action disposition,
and final authority are separate claims. Success at an earlier stage
cannot lend authority to a later stage that has not earned it.

\subsubsection{Which explanations no longer
fit}\label{which-explanations-no-longer-fit}

Two rival explanations do not survive the frozen fourth-study record.

The first is mechanical failure. The unfavorable result was not produced
by a broken run, an incorrect join, a denominator error, missing output
custody, or a silent reconciliation change. Two no-retry replicas
completed. Clean-root reconstruction agreed. The full output-version
census reconciled. Candidate bytes remained frozen. Independent
verification passed {[}9{]}.

The second is retrieval starvation. The lexical router recovered every
governing and blocking policy on the sealed surface. More decisively,
the exhaustive semantic comparator sent every policy to the same
relation model and aggregation rule. It produced the same predictions.
Supplying more policies changed computational burden, but it did not
change a single disposition.

The failure therefore sits downstream of retrieval in this exact
mechanism. The experiment locates the failure. It does not yet explain
why that mechanism failed.

\subsubsection{What remains open}\label{what-remains-open}

The corpus remains open. The fourth study used an authored policy world
designed to expose machine-verifiable ground truth. Its route terms and
controlled language may have made localization easier while leaving
relation interpretation artificial in ways that do not transfer.

The representation remains open. Each proposed action and policy entered
the tested relation stage through whole-text vector representations. The
studies did not determine whether those representations discarded the
distinctions that mattered, or whether a stronger reader could have
recovered them.

The model, encoder, threshold, and aggregation rule remain open. The
fourth study tested one deterministic multinomial logistic relation
reader, two fixed encoders, a selected-model probability threshold of
\texttt{0.70}, and a fixed aggregation rule that discarded
NON\_GOVERNING predictions, escalated an empty remainder or any retained
ambiguous or below-threshold prediction, then returned HOLD for any
retained OPPOSED prediction and ACT otherwise. Changing any one of those
surfaces creates a new scientific question. It does not repair or
reinterpret the frozen result.

Final authority remains open as well. A machine might make the
disposition directly. It might present structured evidence to a
deterministic mechanism. It might abstain. It might preserve a human
decision at the end. The completed studies do not choose among those
designs.

Nor does the failure leave an empty field around a preferred successor.
Structured policy workflows already separate applicability, decision,
and enforcement {[}1, 2{]}. Direct natural-language inference and stance
models can classify an action relative to a supplied reference {[}3,
4{]}. Selective prediction can preserve uncertainty rather than force a
decision. Runtime shields and state machines can constrain actions
through explicit models {[}5{]}. Human and hybrid review remain serious
alternatives when the consequences of a false ACT are high.

\subsubsection{The question the failure
earned}\label{the-question-the-failure-earned}

The system could identify a policy because the action shared language or
semantic proximity with it. That did not establish whether the actor had
the required authority, whether a condition had been met, whether an
exception applied, or which consequence followed. A policy can be
relevant to an action without the action activating the policy's
consequence.

These results led us to revise our hypothesis: judgment in AI agents
requires developing a consequence graph. Such a graph would connect the
actor and authority to policy conditions, exceptions, and the changes an
action would produce.

The studies did not test that structure or establish its necessity. The
result gives us reason to compare it with approaches that do not use an
explicit graph.

Finding the relevant policy and deciding whether it activates are
different problems. The original measurement lineage found enough of the
first problem to expose the second.

\subsection{5. Testing the consequence-graph
hypothesis}\label{testing-the-consequence-graph-hypothesis}

Return to the credential transfer. The authorization, handling, and
confidentiality policies all have something to say about it. Yet the
words \emph{move credential} do not tell us whether the agent is
securing the owner's secret or sending it to someone who has no right to
receive it. Change the destination and recipient, and much of the
language remains the same. The consequence can change completely.

The proposed graph would make these relationships explicit: who is
acting, under what authority, which conditions and exceptions apply, and
what the action would change. The first follow-on study would ask
whether this structure improves relation interpretation. Later studies
would ask whether those interpreted relationships can support reliable
ACT, HOLD, or ESCALATE decisions when policies apply together.

The first follow-on question is therefore specific: \textbf{when the
governing policies are supplied, can a consequence-graph representation
improve relation interpretation over the frozen measurement reader and
serious alternatives?}

\subsubsection{What would count as an
answer}\label{what-would-count-as-an-answer}

The first comparison should hold retrieval fixed. Each candidate would
receive the same governing policies, so a missed relation cannot be
excused by saying the policy was never found. The frozen three-measure
reader provides one baseline. A direct language-based relation reader
offers another: it may be able to infer whether an action satisfies,
opposes, or leaves a policy uncertain without an explicit intermediate
structure {[}3, 4{]}. Where the action and policy can be expressed as
structured facts and predicates, a conventional policy workflow supplies
a third serious comparison {[}1, 2{]}. The proposed consequence graph
would have to earn its place against these alternatives.

The study would need to define relation ground truth, comparison rules,
and meaningful error thresholds before scoring held-out cases. It would
need cases where similar language leads to different answers because
actor, authority, destination, condition, or exception changes. It would
also need to preserve uncertainty. A model that stops escalating by
calling prohibited actions safe has not solved the problem.

An unfavorable result would matter. If the proposed graph adds no
reliable relation utility over the direct reader or the structured
workflow, that approach has not earned a larger claim on the tested
surface. If it works only when policy language is carefully authored,
the result belongs to that construction. If it improves relation
classification but cannot support safe ACT, HOLD, and ESCALATE
decisions, relation and disposition remain separate jobs.

The order matters. First establish whether the system can interpret the
policy's relation to the proposed action. Then ask whether that
interpretation can support a decision to act, hold, or escalate. A
routing gain cannot answer the first question. A relation gain would
not, by itself, answer the second.

Even a graph that outperformed these baselines would establish
comparative utility on that test, not prove that every form of AI
judgment requires a graph.

\subsection{6. Conclusion}\label{conclusion}

We asked whether semantic geometry could give an AI a basis for judgment
before it acts. Across four studies, the tested measurements did not
establish that capability. Raw Euclidean distance retained bounded
proximity information, and cosine retained an encoder-sensitive
directional role. Those signals did not compose into reliable decisions
about whether to act, hold, or seek review.

The final study separated finding the governing policies from
interpreting their relations to an action. On the authored synthetic
surface, a lexical router recovered every governing and blocking policy
while reducing median policy checks by 97.7 percent. The composed
pipeline nevertheless escalated all 2,304 sealed actions in each
encoder. Supplying every policy to the same relation reader and
aggregation rule changed no decision. The failure lay downstream of
retrieval in that mechanism; the studies leave its precise cause open.

These results led us to revise our hypothesis: judgment in AI agents
requires developing a consequence graph. Such a graph would connect the
actor and authority to policy conditions, exceptions, and the changes an
action would produce. In the credential illustration, the agent would
have to connect the owner's authorization with the restrictions on where
the secret can go. Permission to secure a credential does not grant
permission to disclose it to an unauthorized recipient.

We have not yet tested a consequence graph. The graph must earn its
place against direct language readers and structured policy workflows.
Follow-on studies will ask whether making its relationships explicit
improves relation interpretation and, separately, supports reliable
decisions to act, stop, or seek review. The credential case was not part
of the tested corpus. It illustrates the practical distinction we want
the agent to make: complete the transfer that secures the owner's secret
while stopping the one that exposes it.

\subsection{Appendix A. Methods for the fourth
study}\label{appendix-a.-methods-for-the-fourth-study}

The fourth study separated three experiments: routing, relation
interpretation, and their end-to-end composition. The counts below
distinguish the full action-policy world from the sampled relation test
{[}9{]}.

{\def\LTcaptype{none} 
\begin{longtable}[]{@{}lrrr@{}}
\toprule\noalign{}
Population & Calibration & Validation & Sealed \\
\midrule\noalign{}
\endhead
\bottomrule\noalign{}
\endlastfoot
Disjoint mission families & 12 & 6 & 12 \\
Actions, including perturbations & 2,304 & 1,152 & 2,304 \\
Full action-policy pairs & 129,024 & 64,512 & 129,024 \\
\end{longtable}
}

There were 30 families, 5,760 actions, and 322,560 pairs in total. Each
base action had seven perturbations. The routing and composed
experiments each evaluated 2,304 sealed actions per encoder. The
relation experiment instead used 3,840 sampled sealed pairs per encoder,
or 320 per sealed family, containing governing policies and explicit
non-governing controls. These pair rows are not 3,840 independent action
decisions. The twelve sealed families supplied the independent
inferential units.

\subsubsection{Encoders and
measurements}\label{encoders-and-measurements}

The first encoder in the results and figures is
\texttt{sentence-transformers/all-mpnet-base-v2}.

Frozen revision: \texttt{e8c3b32edf5434bc2275fc9bab85f82640a19130}.

The second encoder is \texttt{intfloat/e5-base-v2}.

Frozen revision: \texttt{f52bf8ec8c7124536f0efb74aca902b2995e5bcd}.

Both used attention-masked mean pooling and a maximum length of 512
tokens. The stored vectors were float32 before normalization. The first
encoder used no action or reference prefix. The second prefixed actions
with \texttt{query:} and references with \texttt{passage:}, followed in
each case by one space. Cosine similarity, raw Euclidean distance, and
signed raw dot product remained separate feature columns; unit-vector
normalization was an audit control rather than a replacement for those
raw measurements.

\subsubsection{Fitting, selection, and
evaluation}\label{fitting-selection-and-evaluation}

Each of the seven relation arms used a deterministic multinomial
logistic reader with L2 regularization. Feature means and scales were
fitted on calibration rows only and reused unchanged for evaluation.
Calibration selected \(C \in \{0.1, 1.0, 10.0\}\) by the highest
relation macro-F1, then the lowest multiclass Brier score, then smaller
\(C\). Validation was descriptive. Sealed data selected neither the
model nor the threshold.

The reader predicted ALIGNED, OPPOSED, AMBIGUOUS, and NON\_GOVERNING.
The fixed probability threshold was 0.70. A separate deterministic
aggregation mapped relations to ACT, HOLD, or ESCALATE; no
action-disposition model was fitted. The full relation arm was compared
with its best singleton using the paired effects of the twelve sealed
families, not by treating all pair rows as independent. Governing and
explicit non-governing relation strata remain distinct in the study
evidence.

\subsection{Research transparency}\label{research-transparency}

\subsubsection{Materials availability}\label{materials-availability}

The scientific companion for all four studies is available in the
\href{https://github.com/RadiantManifoldInstitute/semantic-geometry-judgement/releases/tag/v0.1.0}{v0.1.0
release}, with a
\href{https://github.com/RadiantManifoldInstitute/semantic-geometry-judgement/tree/2f62ec3591764476379ec0a09b442d145eabdecc}{fixed
source revision} and
\href{https://github.com/RadiantManifoldInstitute/semantic-geometry-judgement/releases/download/v0.1.0/semantic-geometry-judgement-v0.1.0.tar.gz}{downloadable
archive} {[}6, 7, 8, 9{]}. It contains frozen protocols, scientific
source, synthetic construction contracts, saved results, and available
offline inputs. Readers can inspect the reported outputs and use the
supplied saved-data procedures. The fourth study's saved-summary
inference was independently recalculated at the stored precision; that
check does not reconstruct every primitive row or rerun the models.

The
\href{https://github.com/RadiantManifoldInstitute/semantic-geometry-judgement/blob/2f62ec3591764476379ec0a09b442d145eabdecc/reproduction/README.md}{reproduction
guide} specifies coverage and remaining gaps. Separate row predictions
for some first-study controls are absent. Second- and third-study
procedures use frozen thresholds rather than repeat calibration.
Fourth-study primitive relation recalculation still requires exact gold
and ordering joins; full routing-budget and serial comparisons also
require original feature inputs not supplied in the archive. Model
assets are not bundled, and a complete active dependency environment is
not established for every study. Full experimental replication is
therefore not claimed.

Internal correspondence, administrative approvals, credentials, and
private operational records are excluded. The credential-transfer
example and the task-chain diagrams remain illustrations, not additional
empirical results. The companion's MIT license covers its original
InflectAI contributions; it neither relicenses third-party assets nor
grants rights under the Foundation's patents.

\subsubsection{AI assistance}\label{ai-assistance}

The author originated and directed the research question, conceptual
framework, analysis, and conclusions presented in this work. AI-assisted
tools, including OpenAI Codex, were used under the author's direction to
assist with code and harness preparation, experiment packaging, and
verification workflows. They also assisted with literature organization,
drafting, editing, diagram preparation, and technical formatting. The
named author retains responsibility for the content.

\subsubsection{Competing interests}\label{competing-interests}

The author is the founder of InflectAI and principal investigator at the
Radiant Institute for Manifold Studies. He is an inventor on pending
patent applications concerning reflexic drift measurement and governance
that have been assigned to The Seldon Reflex Foundation, where he also
holds a governing role. InflectAI controls copyright in its original
research-companion contributions; the Foundation holds the related
patent rights. A copyright license for the companion does not grant
permission under the Foundation's patents. This disclosure makes no
claim about patent validity, priority, scope, or scientific efficacy.

\subsection{References}\label{references}

{[}1{]} OASIS. \emph{eXtensible Access Control Markup Language Version
3.0}. OASIS Standard (2013).
\href{https://docs.oasis-open.org/xacml/3.0/xacml-3.0-core-spec-os-en.html}{Standard}.

{[}2{]} Open Policy Agent project. \emph{Open Policy Agent
Documentation}. Accessed 2026-09-08.
\href{https://www.openpolicyagent.org/docs}{Documentation}.

{[}3{]} S. R. Bowman, G. Angeli, C. Potts, and C. D. Manning. ``A Large
Annotated Corpus for Learning Natural Language Inference.''
\emph{Proceedings of EMNLP} (2015).
\href{https://aclanthology.org/D15-1075/}{Paper}.

{[}4{]} S. Mohammad, S. Kiritchenko, P. Sobhani, X. Zhu, and C. Cherry.
``SemEval-2016 Task 6: Detecting Stance in Tweets.'' \emph{Proceedings
of SemEval} (2016). \href{https://aclanthology.org/S16-1003/}{Paper}.

{[}5{]} M. Alshiekh et al.~``Safe Reinforcement Learning via
Shielding.'' \emph{AAAI} 32(1) (2018).
\href{https://doi.org/10.1609/AAAI.V32I1.11797}{Paper}.

\subsection{Research evidence and replication
materials}\label{research-evidence-and-replication-materials}

The following study packages are included in the v0.1.0 scientific
companion. Each link below points to the fixed source revision used for
this paper.

{[}6{]} Radiant Institute for Manifold Studies. \emph{Study 1: Compact
representation and relation recovery}. Scientific companion v0.1.0
(2026).
\href{https://github.com/RadiantManifoldInstitute/semantic-geometry-judgement/tree/2f62ec3591764476379ec0a09b442d145eabdecc/studies/01-scalar-and-norms}{Methods
and protocol}.
\href{https://github.com/RadiantManifoldInstitute/semantic-geometry-judgement/blob/2f62ec3591764476379ec0a09b442d145eabdecc/results/C/scientific-result.json}{Saved
results}.

{[}7{]} Radiant Institute for Manifold Studies. \emph{Study 2:
Proximity, direction, and consequence measurements}. Scientific
companion v0.1.0 (2026).
\href{https://github.com/RadiantManifoldInstitute/semantic-geometry-judgement/tree/2f62ec3591764476379ec0a09b442d145eabdecc/studies/02-component-measures}{Methods
and protocol}.
\href{https://github.com/RadiantManifoldInstitute/semantic-geometry-judgement/blob/2f62ec3591764476379ec0a09b442d145eabdecc/results/D/scientific-result.json}{Saved
results}.

{[}8{]} Radiant Institute for Manifold Studies. \emph{Study 3: Composed
routing and action decisions}. Scientific companion v0.1.0 (2026).
\href{https://github.com/RadiantManifoldInstitute/semantic-geometry-judgement/tree/2f62ec3591764476379ec0a09b442d145eabdecc/studies/03-two-stage-policy-checks}{Methods
and protocol}.
\href{https://github.com/RadiantManifoldInstitute/semantic-geometry-judgement/blob/2f62ec3591764476379ec0a09b442d145eabdecc/results/E/scientific-result.json}{Saved
results}.

{[}9{]} Radiant Institute for Manifold Studies. \emph{Study 4:
Separating policy retrieval, relation interpretation, and action
disposition}. Scientific companion v0.1.0 (2026).
\href{https://github.com/RadiantManifoldInstitute/semantic-geometry-judgement/tree/2f62ec3591764476379ec0a09b442d145eabdecc/studies/04-routing-relation-disposition}{Methods
and protocol}.
\href{https://github.com/RadiantManifoldInstitute/semantic-geometry-judgement/blob/2f62ec3591764476379ec0a09b442d145eabdecc/results/F/scientific-result.json}{Saved
results}.

\end{document}